\documentclass[pdflatex,sn-chicago]{sn-jnl}%
\usepackage[utf8]{inputenc}%
\usepackage{booktabs, longtable, graphicx, caption, pdflscape}%
\usepackage{multirow}%
\usepackage{amsmath,amssymb,amsfonts}%
\usepackage{amsthm}%
\usepackage[title]{appendix}%
\usepackage{xcolor}%
\usepackage{algorithm}%
\usepackage{algorithmicx}%
\usepackage{algpseudocode}%
\theoremstyle{thmstyleone}%
\theoremstyle{thmstyletwo}%
\theoremstyle{thmstylethree}%
\usepackage{adjustbox}%
\usepackage{geometry}%
\usepackage{array}%
\DeclareUnicodeCharacter{2212}{-}%
\begin{document}%

\title{Regional climate projections using a deep-learning–based model-ranking and downscaling framework: Application to European climate zones}%

\author*[1]{\fnm{Parthiban} \sur{Loganathan}}\email{parthi@kth.se}
\author[1]{\fnm{Elias} \sur{Zea}}
\author[1]{\fnm{Ricardo} \sur{Vinuesa}}
\author[1]{\fnm{Evelyn} \sur{Otero}}
\affil[1]{\orgdiv{Department of Engineering Mechanics}, \orgname{KTH Royal Institute of Technology}, \orgaddress{\city{Stockholm}, \country{Sweden}}}
\maketitle

\section*{Abstract}

Accurate regional climate forecast calls for high-resolution downscaling of Global Climate Models (GCMs). This work presents a deep-learning-based multi-model evaluation and downscaling framework ranking 32 Coupled Model Intercomparison Project Phase 6 (CMIP6) models using a Deep Learning-TOPSIS (DL-TOPSIS) mechanism and so refines outputs using advanced deep-learning models. Using nine performance criteria, five Köppen-Geiger climate zones—Tropical, Arid, Temperate, Continental, and Polar—are investigated over four seasons. While TaiESM1 and CMCC-CM2-SR5 show notable biases, ranking results show that NorESM2-LM, GISS-E2-1-G, and HadGEM3-GC31-LL outperform other models. Four models contribute to downscaling the top-ranked GCMs to 0.1° resolution. Vision Transformer (ViT), Geospatial Spatiotemporal Transformer with Attention and Imbalance-Aware Network (GeoSTANet), CNN-LSTM, CNN-Long Short-Term Memory (ConvLSTM). Effectively capturing temperature extremes (TXx, TNn), GeoSTANet achieves the highest accuracy (Root Mean Square Error (RMSE) = 1.57°C, Kling-Gupta Efficiency (KGE) = 0.89, Nash-Sutcliffe Efficiency (NSE) = 0.85, Correlation (r) = 0.92, so reducing RMSE by 20\% over ConvLSTM. CNN-LSTM and ConvLSTM do well in Continental and Temperate zones; ViT finds fine-scale temperature fluctuations difficult. These results confirm that multi-criteria ranking improves GCM selection for regional climate studies and transformer-based downscaling exceeds conventional deep-learning methods. This framework offers a scalable method to enhance high-resolution climate projections, benefiting impact assessments and adaptation plans.

\textbf{Keywords} Climate Downscaling, GCM Ranking, High-Resolution Climate Projections, Köppen-Geiger Climate Zones, Regional Climate Impact, Transformer-Based Downscaling.

\section{Introduction}

Considered one of the most pressing global concerns of the twenty-first century, climate change has possibly major consequences on ecosystems, infrastructure, and society all around \cite{1}. Understanding and forecasting climate dynamics has never been more important as world temperatures rise and severe storms become more frequent. Expanding our scientific knowledge of past, current, and future climatic conditions in this setting depends on numerical models that faithfully depict the intricate physical processes of the Earth system. Among climate modelling agencies, the Coupled Model Intercomparison Project (CMIP) has become a core cooperative effort promoting worldwide cooperation and standardizing GCMs \cite{2}. Currently in its sixth phase (CMIP6), the project has driven the creation of ever more advanced GCMs simulating climate variability and change with improved realism \cite{3}. These models usually run at coarse resolutions—often between 50 and 250 km grid cells—that are insufficient for resolving localized climate events, even if they provide insightful analysis at global and continental scales. Such coarse resolution reduces the capacity to represent important regional aspects including localized precipitation extremes, temperature changes in complicated topography, and subtle land-sea contrasts \cite{4,5}. Effective regional climate impact assessments and adaptation planning depend on downscaling techniques that can convert coarse GCM data into high-resolution localized estimates, so there is a great demand for them.

Emerging as a vital answer to the gap between the coarse spatial resolution of GCM outputs and the high-resolution data needed for local decision-making are downsizing methods. Dynamic and statistical downscaling are the two main approaches that are now in use. Dynamic downscaling nests high-resolution simulations inside the boundary conditions given by global models using Regional Climate Models (RCMs). This method enables the explicit simulation of local physical processes, including those affected by intricate topography and land-use features \cite{6,7}. Statistical downscaling techniques, on the other hand, depend on experimentally generated connections between local observations and major climate factors. Fine-resolution climate data has been produced using extensively applied techniques like quantile mapping, weather typing, weather generators, and regression-based processes. Generally speaking, statistical approaches are preferred because of their reduced computational requirements and capacity to include site-specific observational data—which can be very helpful when addressing localized climate events \cite{8}. Both dynamic and statistical downscaling techniques have difficulties notwithstanding their strengths, especially in areas with high climatic fluctuation where observational data may be few. The need to choose the most appropriate GCMs for downscaling becomes even more evident as regional studies progressively need customized climate projections, enabling sophisticated model evaluation and selection methods.

Choosing the best GCM for regional climate research is a difficult and important chore since different models may show differing performances depending on the certain region and climate variable under examination. Variability in model physics, parameterizations, initial circumstances, and the modelling of important feedback processes means no single model can routinely outperform others across all scenarios \cite{9,10}. Multi-criteria decision-making (MCDM) techniques provide a disciplined framework for assessing several, occasionally contradictory, performance indicators to negotiate this complexity. The Technique for Order Preference by Similarity to Ideal Solution (TOPSIS), which ranks options by evaluating their proximity to an ideal solution while concurrently calculating their distance from the anti-ideal, or worst-case scenario \cite{11,12}, is one well-established MCDM method. Conversely, conventional TOPSIS implementations depend on either fixed or subjectively set weights for every performance criterion, a method that could result in biased or inconsistent model selections \cite{13}. DL-TOPSIS has been created as a novel approach to solve this restriction. DL-TOPSIS offers a more objective and flexible ranking system by learning a neural network to determine the ideal weights straight from the data. This method combines a wide range of performance criteria—including bias, Root Mean Square Error (RMSE), Pearson correlation (r), Kling-Gupta Efficiency (KGE), Nash-Sutcliffe Efficiency (NSE), and measures of distribution overlap—to guarantee that the chosen GCM is best suited for catching the climatic characteristics relevant to a given region \cite{14,15}.

Defined as the variation between the daily maximum (tasmax) and minimum (tasmin) temperatures, the diurnal temperature range (DTR) has become more important as a climatic indicator with broad effects on agricultural output, human health, and ecological systems \cite{16,17}. A significant indicator of underlying changes in air circulation, cloud cover, and land–surface interactions, variations in DTR can reflect \cite{18,19} climatic dynamics. Downscaling attempts historically concentrated on mean temperature or precipitation, but increasing attention is being paid to precisely capturing the DTR. Most statistical early downscaling strategies use quantile mapping, weather typing, and linear regression. While these methods are computationally efficient, they frequently fail to adequately depict non-linear connections and intricate inter-dependencies between climatic variables \cite{20}.

Deep learning has transformed pattern identification in big datasets in recent years thanks to early applications modelling correlations between coarse GCM outputs and local observations using multi-layer perceptrons (MLPs) \cite{21,22}. Recent developments in deep learning methods spanning from convolutional neural networks to transformer-based architectures have shown a strong ability for modelling and predicting challenging physical events in atmospheric modelling \cite{23,24,25}. MLPs could not, however, efficiently use spatial information included in climate data. By allowing the extraction of spatial information capturing localized meteorological events, Convolutional Neural Networks (CNNs) offered a major improvement. Recurrent Neural Networks (RNNs) and Long Short-Term Memory (LSTM) networks have been used concurrently to model temporal dependencies, such as seasonal cycles and sequential variability, so improving the capacity to replicate daily and seasonal oscillations in DTR \cite{26,27}. To handle the complexity of downscaling DTR \cite{28,29}, hybrid models combining CNNs with LSTMs, as well as methods using Generative Adversarial Networks (GANs), have shown great potential.

Building on these advances, transformer-based deep-learning models have brought fresh capabilities for climate downscaling. Representing the forefront in modelling spatiotemporal climate data, ViTs and the specialized GeoSTANet capture long-range spatial dependencies using self-attention mechanisms, therefore allowing the modelling of broad atmospheric circulation patterns. Though ViTs show great promise, they can find it challenging to generalize across areas with quite different climate conditions. Conversely, GeoSTANet is especially meant to solve these difficulties by combining transformer-based attention mechanisms with an imbalance-aware training approach, therefore enabling it to thrive in both catching extreme occurrences and regular climate patterns. In recreating fine-scale characteristics, diurnal cycles, and seasonal transitions, these advanced hybrid architectures—including CNN-LSTM and ConvLSTM models—have been demonstrated to be quite superior to conventional statistical methods \cite{30}. Notwithstanding their remarkable performance, these deep-learning techniques have several difficulties including the need for large, high-quality datasets like ERA5 data \cite{31}, rising computational demands connected with high-resolution spatiotemporal grids \cite{32}, possible over-fitting risk, and problems with model interpretability \cite{33,34}. Reliable support of advanced downscaling models for regional climate impact assessments depends on addressing these problems.

 The present study defines five climate zones (Tropical, Arid, Temperate, Continental, and Polar) reflecting the approximate climatological variations across the region, each characterized by unique temperature and precipitation patterns influencing GCM performance and the representation of extremes \cite{35,36}. The present work's core argument is that the initial choice of the best-performing GCMs for a specific region determines optimal downscaling performance  \cite{37,38,39}. First, the DL-TOPSIS methodology is used to objectively rank CMIP6 models based on a comprehensive set of performance metrics including extreme temperature indicators maximum tasmax (TXx) and lowest tasmin (TNn), Probability Density Function overlap (PDF Overlap), standard deviation differences (SD Diff) as well as traditional metrics like bias, RMSE, correlation, NSE, and KGE \cite{40,41}. Advanced deep-learning downscaling methods including CNN-LSTM, ConvLSTM, ViT, and GeoSTANet are then fed, top-ranked models. Essential for climate services and impact assessments in sectors including agriculture, urban planning, health, and energy, this combined approach not only reduces the transmission of biases from coarse-scale models but also improves the accuracy of high-resolution forecasts \cite{42,43}. Moreover, by using high-performance computational resources and large-scale observational data, the proposed methodology offers a scalable and repeatable blueprint for closing the gap between local adaptation techniques and global climate modelling \cite{44,45}. In the end, this multidisciplinary approach merging advanced machine learning with operations research has the potential to produce more accurate and useful climate projections for practitioners and legislators all around.
 
\section{Study area and Data description}

\begin{figure}[h]
    \centering
    \includegraphics[width=0.9\textwidth]{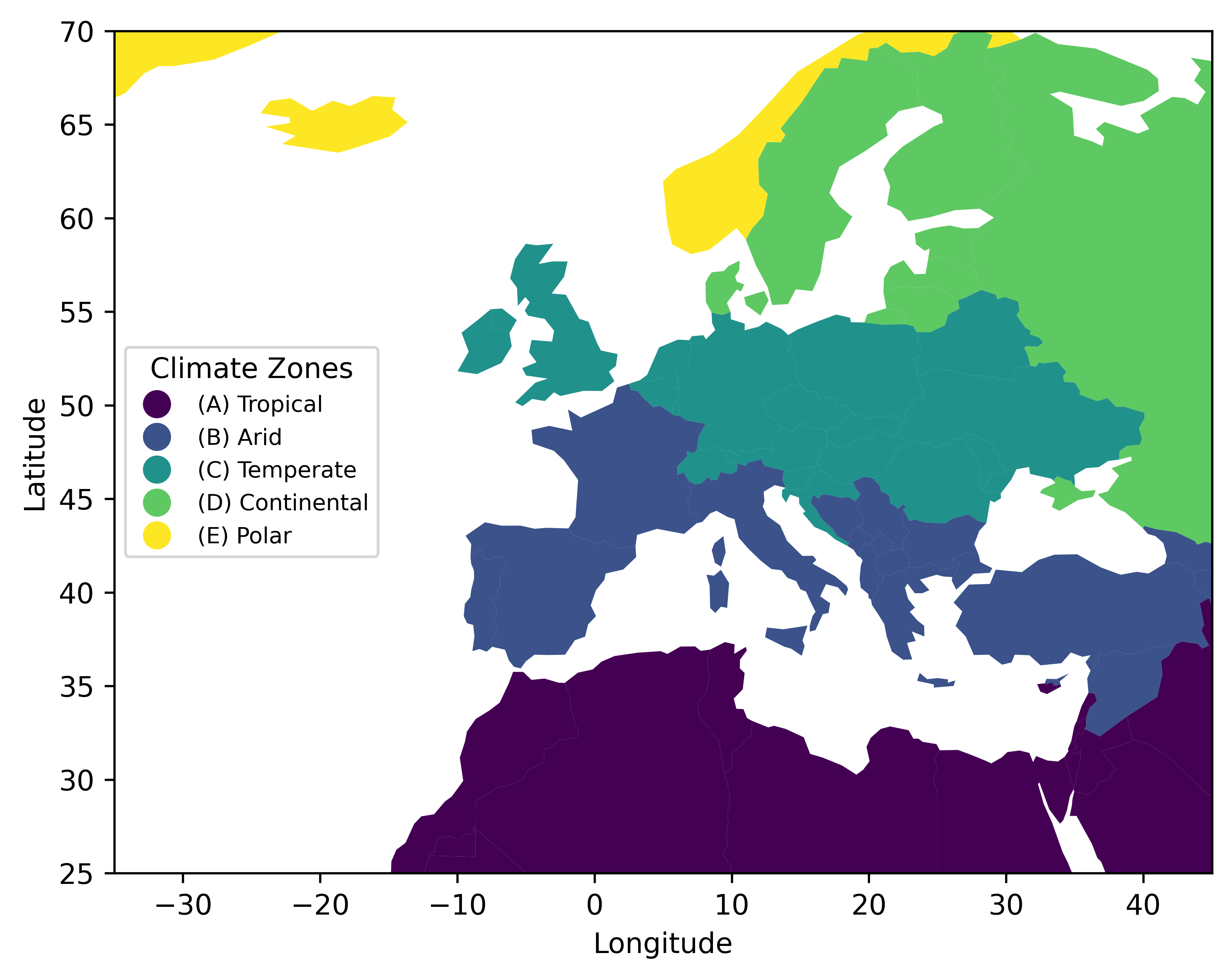}
    \caption{Spatial representation of Köppen-Geiger climate zones classification across Europe.}
    \label{fig:1}
\end{figure}

Advanced downscaling methods find a useful proving ground in Europe's varied climatic zones, which range from Mediterranean beaches and temperate coastal areas to continental interiors and Arctic tundra \cite{46,47}. From the warm Mediterranean basin to the Arctic tundra, Europe offers one of the most climatically varied areas worldwide and is thus a perfect testbed for assessing climate models and downscaling methods. Based on temperature and precipitation patterns, the Köppen-Geiger climate classification offers a disciplined approach to split the continent into five main zones: Tropical, Arid, Temperate, Continental, and Polar. These categories allow one to evaluate model performance geographically with a customizing effect. Different seasonal changes in every zone affect local temperature extremes, air circulation patterns, and precipitation dynamics. The North Atlantic Oscillation (NAO) significantly affects Winter, thereby influencing storm courses and temperature variation. Spring marks a change in temperature and a changing precipitation zone phase. Heat waves and convective activity define Summer, especially in Temperate and Arid zones; Autumn marks a resumption of baroclinic activity and mid-latitude storm systems. Evaluating model performance depends on these seasonal dynamics as various GCMs might struggle with certain regional climatic variables. Figure \ref{fig:1} shows the Köppen-Geiger classification for Europe, stressing the spatial distribution of the climatic zones applied in this work.

Thirty-two CMIP6 models were chosen to evaluate climate model accuracy using a variety of geographical resolutions (50–250 km) and many parameterizing approaches. Coordinated by the World Climate Research Program (WCRP), the CMIP6 repository offers the most developed suite of GCMs accessible for historical and future climate simulations. Different physical parameterization, land-atmosphere coupling, and climate sensitivity among these models affect their capacity to replicate regional climatic variability. Data availability, the inclusion of tasmax and tasmin, and the adoption of a standardized r1i1p1f1 ensemble member to provide consistent initialization-guided model selection. Developed by ECMWF, observational benchmarks originate from ERA5 data and offer high-resolution data at 0.1° x 0.1° ($\sim$10km). ERA5 produces continuous, high-quality climate records by combining satellite, in situ, and reanalysis data, unlike crude GCM outputs. Selected to provide consistent assessment between CMIP6 historical simulations and observational data was the historical period 1985–2014. Table \ref{tab:1} gives a summary of the thirty-two CMIP6 GCMs together with their geographic resolution, original institution, and salient features.

\begin{landscape}
\begin{table}[h]
\centering
\renewcommand{\arraystretch}{1}
\scriptsize
\begin{longtable}{c l l l p{8.5cm}}
    \caption{List of CMIP6 models used in the study.}
    \label{tab:1} \\
    \toprule
    \textbf{S. No} & \textbf{Model} & \textbf{Institute (Country)} & \textbf{Resolution (km)} & \textbf{Key Features} \\
    \midrule
    \endfirsthead
    \bottomrule
    \endlastfoot
    1  & ACCESS-CM2        & CSIRO (Australia)                     & $\sim$100 & Improved ocean-atmosphere coupling \\
    2  & ACCESS-ESM1-5     & CSIRO (Australia)                     & $\sim$100 & Earth System Model with Advanced Carbon Cycle \\
    3  & BCC-CSM2-MR       & Beijing Climate Center (China)        & $\sim$100 & Medium resolution; aerosol-cloud interactions \\
    4  & BCC-ESM1          & Beijing Climate Center (China)        & $\sim$280 & Coupled Earth System Model with biogeochemistry \\
    5  & CAMS-CSM1-0       & Chinese Academy of Meteorological Sciences & $\sim$100 & Atmospheric physics enhancements \\
    6  & CAS-ESM2-0        & Chinese Academy of Sciences (China)   & $\sim$100 & Integrated land-vegetation-atmosphere coupling \\
    7  & CESM2             & NCAR (USA)                            & $\sim$110 & Biogeochemical cycles; high-resolution processes \\
    8  & CESM2-FV2         & NCAR (USA)                            & $\sim$50  & A high-resolution version of CESM2 \\
    9  & CESM2-WACCM       & NCAR (USA)                            & $\sim$110 & Whole Atmosphere Coupling; upper-atmosphere focus \\
    10 & CIESM             & Chinese Institute of Earth System Modeling & $\sim$100 & Improved monsoon simulations \\
    11 & CMCC-CM2-SR5      & CMCC (Italy)                          & $\sim$100 & High-resolution atmosphere-ocean coupling \\
    12 & CMCC-ESM2         & CMCC (Italy)                          & $\sim$100 & Dynamic vegetation and Earth system processes \\
    13 & CNRM-CM6-1        & CNRM (France)                         & $\sim$100 & Advanced surface energy balance representation \\
    14 & CNRM-ESM2-1       & CNRM (France)                         & $\sim$100 & Earth System Model with carbon-climate feedbacks \\
    15 & CanESM5           & CCCma (Canada)                        & $\sim$250 & Enhanced aerosol-radiation interactions \\
    16 & EC-Earth3         & EC-Earth Consortium (Europe)          & $\sim$100 & Focus on European climate variability \\
    17 & FGOALS-f3-L       & IAP (China)                           & $\sim$100 & Precipitation physics improvement \\
    18 & FGOALS-g3         & IAP (China)                           & $\sim$280 & Global hydrological cycle representation \\
    19 & GFDL-CM4          & GFDL (USA)                            & $\sim$100 & High-resolution ocean-atmosphere coupling \\
    20 & GFDL-ESM4         & GFDL (USA)                            & $\sim$100 & Earth System Model with advanced biogeochemistry \\
    21 & INM-CM4-8         & INM (Russia)                          & $\sim$150 & Coupled dynamical processes \\
    22 & INM-CM5-0         & INM (Russia)                          & $\sim$150 & Improved cloud parametrization \\
    23 & IPSL-CM6A-LR      & IPSL (France)                         & $\sim$250 & Advanced aerosol-cloud interactions \\
    24 & KACE-1-0-G        & KMA (Korea)                           & $\sim$100 & Focus on regional climate dynamics \\
    25 & MIROC6            & MIROC (Japan)                         & $\sim$100 & Multi-scale atmosphere-ocean coupling \\
    26 & MPI-ESM1-2-HR     & MPI-M (Germany)                       & $\sim$100 & A high-resolution version of MPI Earth System Model \\
    27 & MPI-ESM1-2-LR     & MPI-M (Germany)                       & $\sim$250 & Low-resolution Earth System Model \\
    28 & MRI-ESM2-0        & MRI (Japan)                           & $\sim$100 & Ocean biogeochemistry and carbon dynamics \\
    29 & NESM3             & NUIST (China)                         & $\sim$100 & Improved monsoon and hydrological cycles \\
    30 & NorESM2-LM        & NCC (Norway)                          & $\sim$250 & Low-resolution Nordic Earth System Model \\
    31 & NorESM2-MM        & NCC (Norway)                          & $\sim$100 & Medium-resolution Nordic Earth System Model \\
    32 & UKESM1-0-LL       & Met Office (UK)                       & $\sim$100 & Coupled Earth System Model with land-atmosphere focus \\
    \bottomrule
\end{longtable}
\end{table}
\end{landscape}

Performance measures are computed at a constant 0.1° × 0.1° spatial grid for complete model evaluation, guaranteeing a direct comparison between ERA5 and regridded CMIP6 results. Calculated from tasmax and tasmin, the DTR is a fundamental indication of radiative and land-surface processes. Multiple statistical tests are run to measure model biases, including Bias, RMSE, r, KGE, NSE, and PDF Overlap. These measurements evaluate multi-dimensional model quality by capturing mean biases, variability, distributional integrity, and extreme event representation. Seasonal stratification and zone-based categorization let one thoroughly examine model strengths and shortcomings in several climatic regimes. Following downscaling studies will apply the processed datasets and assessment methodology to guarantee that high-resolution climate forecasts are produced from the best-performing models.

\section{Methodology}

This study proposes a hybrid DL-TOPSIS framework to rank CMIP6 climate models based on their historical performance and applies advanced deep-learning architectures for statistical downscaling. The methodological workflow consists of three primary blocks, as presented in Figure \ref{fig:2}:

\begin{figure}[h]
    \centering
    \includegraphics[width=0.95\textwidth]{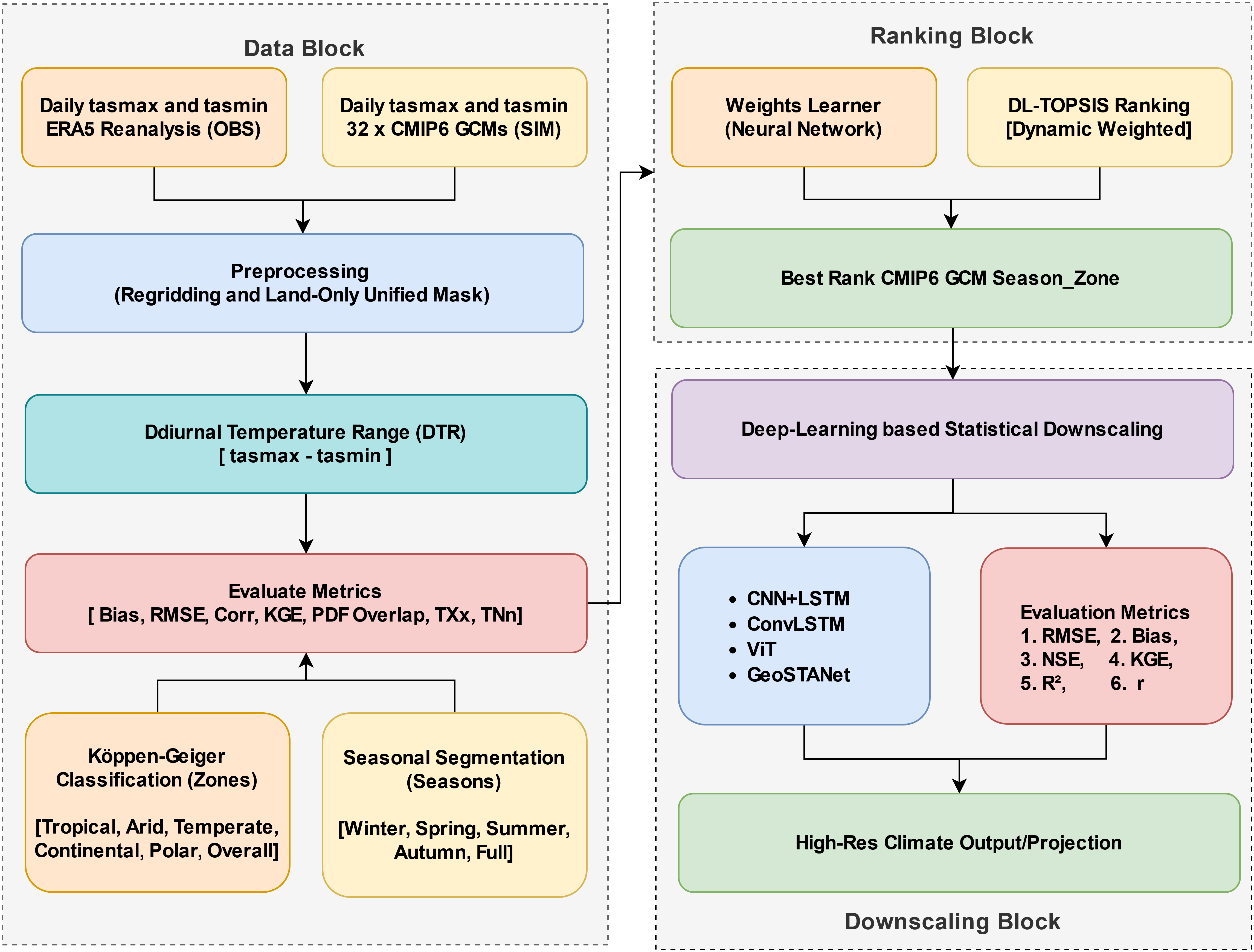}
    \caption{Methodology workflow for model ranking and downscaling.}
    \label{fig:2}
\end{figure}

\begin{itemize}
    \item \textbf{Data Block}: Daily maximum and minimum temperature data (\texttt{tasmax, tasmin}) from ERA5 reanalysis (observations) and 32 CMIP6 models (simulations) are pre-processed.
    \item \textbf{Ranking Block}: The top CMIP6 models are identified using a DL-TOPSIS ranking system, which dynamically assigns weights to performance metrics.
    \item \textbf{Downscaling Block}: The selected top GCM models are downscaled to high-resolution climate projections using deep-learning architectures such as CNN-LSTM, ConvLSTM, ViT, and GeoSTANet.
\end{itemize}

To ensure a strong model assessment, both ranking and downscaling steps use an independent set of performance measures. The proposed framework improves regional climate projections for adaptation and mitigating strategies and allows a fair evaluation of climate model integrity.

\subsection{Data pre-processing block:} The data pre-processing stage ensures that ERA5 and CMIP6 datasets are aligned for direct comparison and statistical downscaling. This process involves multiple steps:
\begin{enumerate}
\item {Data acquisition:} a) Observational data: ERA5 reanalysis dataset (0.1° $\times$ 0.1° resolution, 1985–2014), and b) Climate model simulations: 32 CMIP6 models, historical experiments (r1i1p1f1 ensemble), daily \texttt{tasmax} and \texttt{tasmin}.
\item{Regridding and land-only masking:} CMIP6 models have varying spatial resolutions (50 km to 250 km). To ensure smooth processing, all CMIP6 GCM outputs are scaled using bilinear interpolation to match the ERA5 grid (0.1° $\times$ 0.1°), and a land-only mask is applied to exclude oceanic regions to maintain consistency in model evaluation.
\item{Computation of the DTR:} DTR is derived from \texttt{tasmax} and \texttt{tasmin} for both ERA5 and CMIP6 datasets:
\begin{equation}\text{DTR} = \text{tasmax} - \text{tasmin}\end{equation}
DTR is an important climate analysis metric, representing the day-night temperature contrast and serving as a regional climate variability indicator.
\item{Seasonal and climate zone classification:} For climate-specific assessment, the data is categorized according to: a) \textbf{Seasons}: 1) Winter (December, January, and February), 2) Spring (March, April, and May), 3) Summer (June, July, and August), 4) Autumn (September, October, and November), and 5) Annual; b) \textbf{Climate Zones}: The Köppen-Geiger classification is used to classify data into six zones: 1) Tropical, 2) Arid, 3) Temperate, 4) Continental, 5) Polar, and 6) Entire Europe.
\item{Calculation of evaluation metrics: } A comprehensive set of statistical and physical performance metrics is computed to assess the fidelity of each CMIP6 model: a) Bias, b) RMSE, c) KGE, d) NSE, e) r, f) PDF Overlap, g) Extreme temperature indices: TXx and TNn. The pre-processing steps standardize the ERA5 and CMIP6 datasets, facilitating impartial model evaluation and efficient statistical downscaling in later phases. The performance evaluation metrics and their descriptors are shown in Table \ref{tab:2}.

\end{enumerate}
\begin{landscape}
\scriptsize
\newcolumntype{C}[1]{>{\centering\arraybackslash}p{#1}}
\renewcommand{\arraystretch}{1.2}
\setlength{\extrarowheight}{13pt}
\begin{longtable}{@{}p{6.5cm} C{7cm} p{11cm}@{}}
\caption{Performance metrics for CMIP6 model evaluation.} \label{tab:2} \\
\toprule
\textbf{Parameter} & \textbf{Formula} & \textbf{Significance} \\ 
\midrule
\endfirsthead
\multicolumn{3}{c}{\textit{Table \ref{tab:2} continued from previous page}} \\
\toprule
\textbf{Parameter} & \textbf{Formula} & \textbf{Significance} \\ 
\midrule
\endhead
\midrule
\multicolumn{3}{r}{\textit{Continued on next page}} \\
\midrule
\endfoot
\bottomrule
\endlastfoot
RMSE (Root Mean Square Error) & 
$\displaystyle \sqrt{\frac{1}{n}\sum_{i=1}^{n}(M_i - O_i)^2}$ & Measures overall deviation between predictions and observations; lower is better. \\[2mm]
Bias & 
$\displaystyle \frac{1}{n}\sum_{i=1}^{n}(M_i - O_i)$ & Quantifies systematic error; values near zero indicate minimal bias. \\[2mm]
NSE (Nash-Sutcliffe Efficiency) & 
$\displaystyle 1 - \frac{\sum_{i=1}^{n}(O_i - M_i)^2}{\sum_{i=1}^{n}(O_i - \bar{O})^2}$ & Evaluates predictive skill; values close to 1 indicate high performance. \\[2mm]
KGE (Kling-Gupta Efficiency) & 
\parbox[c]{8cm}{\centering \scriptsize
$1-\sqrt{(r-1)^2+(\beta-1)^2+(\gamma-1)^2}$\\[1mm]
$\beta=\mu_M/\mu_O,\quad \gamma=(\sigma_M/\mu_M)/(\sigma_O/\mu_O)$} & Combines correlation, bias, and variability; optimal value is 1. \\[2mm]
$r^2$ (Coefficient of Determination) & 
$\displaystyle \left(\frac{\sum_{i=1}^{n}(M_i-\bar{M})(O_i-\bar{O})}{\sqrt{\sum_{i=1}^{n}(M_i-\bar{M})^2\,\sum_{i=1}^{n}(O_i-\bar{O})^2}}\right)^2$ & Proportion of variance in observations explained by the model; higher values indicate better performance. \\[2mm]
$r$ (Correlation Coefficient) & 
$\displaystyle \frac{\sum_{i=1}^{n}(M_i-\bar{M})(O_i-\bar{O})}{\sqrt{\sum_{i=1}^{n}(M_i-\bar{M})^2\,\sum_{i=1}^{n}(O_i-\bar{O})^2}}$ & Measures linear association; values near 1 indicate strong positive correlation. \\[2mm]
PDF (Probability Density Function Overlap) & 
$\displaystyle \int \min\Bigl(P_M(x), P_O(x)\Bigr){\rm d}x$ & Quantifies similarity between model and observed distributions; 1 indicates perfect overlap. \\[2mm]
TXx (Maximum Temperature) & 
$\displaystyle \max\bigl(\texttt{tasmax}\bigr)$ & Indicates extreme high temperatures; critical for assessing heat events. \\[2mm]
TNn (Minimum Temperature) & 
$\displaystyle \min\bigl(\texttt{tasmin}\bigr)$ & Indicates extreme low temperatures; essential for evaluating cold events. \\
\end{longtable}
\end{landscape}

\subsection{Model ranking block:}

A hybrid DL-TOPSIS architecture objectively ranks CMIP6 models throughout several climate zones and seasons. This method combines multi-criteria decision-making (MCDM) methodologies with a neural network-based dynamic weighting mechanism to guarantee a strong and data-driven ranking of models. Through their proximity to an ideal solution and distance from an anti-ideal solution, the Technique for Order Preference by Similarity to Ideal Solution (TOPSIS) ranks models. This guarantees that the models with maximum performance also minimize bias and maximize correlation with observable data. The TOPSIS rating comprises five steps:
\begin{enumerate}
\item {Normalization of performance metrics:} Each performance metric $C_{ij}$ for model $i$ and metric $j$ is normalized to remove scale variations:
\begin{equation}N_{ij} = \frac{C_{ij}}{\sqrt{\sum_{i=1}^{m} C_{ij}^2}}\end{equation}
where, $C_{ij}$ is the computed value of metric $j$ for model $i$, and $m$ is the total number of CMIP6 models.
\item{Weighted normalization:} Each normalized metric is multiplied by its respective weight $w_j$, which is learned dynamically using a neural network:
\begin{equation}W_{ij} = w_j \cdot N_{ij}\end{equation}
\item{Calculation of ideal and anti-ideal solutions:} The ideal solution represents the best possible values across all models, while the anti-ideal solution represents the worst:
\begin{equation}A^+ = \{ \max(W_{ij}), \text{ for benefit metrics}; \min(W_{ij}), \text{ for cost metrics} \},\end{equation}
\begin{equation}A^- = \{ \min(W_{ij}), \text{ for benefit metrics}; \max(W_{ij}), \text{ for cost metrics} \},\end{equation}
where, the higher the benefit metrics (KGE, NSE, correlation, and PDF overlap) the better, and the lower the cost metrics (Bias, RMSE) the better.
\item{Euclidean distance calculation:} The Euclidean distance of each model from the ideal and anti-ideal solutions is computed:
\begin{equation}D_i^+ = \sqrt{\sum_{j=1}^{n} (W_{ij} - A_j^+)^2}, \quad D_i^- = \sqrt{\sum_{j=1}^{n} (W_{ij} - A_j^-)^2}\end{equation}
where, $ D_i^+ $ represents the distance from the ideal solution, and $ D_i^- $ represents the distance from the anti-ideal solution.
\item{Computation of Closeness Coefficient:} The closeness coefficient (CC) determines the final ranking of each model:
\begin{equation}CC_i = \frac{D_i^-}{D_i^+ + D_i^-}\end{equation}
where a higher $ CC_i $ value indicates a better-performing model. Models are ranked in descending order of $ CC_i $.
\end{enumerate}
\subsubsection{Dynamic weighting with deep-learning:} For every performance evaluation parameter, traditional TOPSIS implementations used fixed, subjective weights. Whereas, climate model evaluation requires a data-driven approach, where the importance of certain criteria varies based on region and season. A deep neural network (DNN) that dynamically learns ideal metric weights to handle this is presented.
\subsubsection{Neural network architecture}
The neural network is meant to decide ideal ranking metric weights. Its design comprises:
\begin{itemize}
    \item \textbf{Input Layer}: Performance metrics of each model such as Bias, RMSE, KGE, Correlation, etc.
    \item \textbf{Hidden Layers}: Two fully connected layers with 64 and 32 neurons, ReLU activation.
    \item \textbf{Output Layer}: Softmax activation with normalized weights output for each metric.
\end{itemize}

\subsubsection{Neural network training setup}
Minimizing the reconstruction error between expected and observed ranking patterns helps the neural network discover the ideal weighting scheme. The loss function is:
\begin{equation} \text{Loss} = \sum_{i=1}^{n} (W_{ij} - \hat{W}_{ij})^2 \end{equation}
where, $ W_{ij} $ represents the ground-truth weight of metric $ j $ for model $ i $, and $ \hat{W}_{ij} $ is the predicted weight.
    \begin{itemize}
        \item \textbf{Optimizer}: Adam
        \item \textbf{Learning Rate}: 0.001
        \item \textbf{Batch Size}: 32
        \item \textbf{Epochs}: 50
        \item \textbf{Loss Function}: Mean Squared Error (MSE)
    \end{itemize}
The model is trained using stochastic gradient descent (SGD) and then refined using an Adam optimizer \cite{23}. Once trained, the learned weights derived from the neural network replace the TOPSIS system's static weights. This implies that the ranking process is adaptable enough for regional climate conditions, in which multiple performance criteria could have different relevance. The top-ranked models are selected for every season and climate zone after the computation of closeness coefficients ($CC_i$) for every 32 CMIP6 model. The best-performing models then go to the downscaling block, where statistical downscaling techniques rooted in deep-learning are applied. This hybrid DL-TOPSIS approach ensures that high-resolution climate projections only rely on the most trustworthy CMIP6 models, hence improving the accuracy of the next climate assessments.

\subsection{Statistical downscaling block}

This section describes the deep-learning architectures used to downscale coarse-resolution CMIP6 outputs (50–250 km) to a high-resolution grid (0.1$^\circ\times$0.1$^\circ$). Four models are developed to capture both spatial and temporal features: (i) CNN-LSTM, (ii) ConvLSTM, (iii) ViT, and (iv) GeoSTANet.

\begin{enumerate}
\item \textbf{CNN-LSTM:} This model combines CNNs for spatial feature extraction and LSTM networks for temporal sequence modelling. The architecture is described in what follows.
\begin{enumerate}
    \item \textbf{Input:} A climate data cube 
    \begin{equation}X \in \mathbb{R}^{t \times h \times w \times c}\end{equation}
    where $t$ is the number of time steps, $h \times w$ are the spatial dimensions, and $c$ is the number of variables.   
    \item \textbf{CNN block:} Spatial features are extracted through a series of convolutional layers. For the \(l^\text{th}\) layer, the feature map is computed as:
    \begin{equation} F^{(l)}_{ij} = \text{ReLU}\left(\sum_{k=-p}^{p}\sum_{m=-p}^{p} W_{km}^{(l)}\, X^{(l-1)}_{(i+k)(j+m)} + b^{(l)}\right) \end{equation}
    Where, \(W_{km}^{(l)}\) and \(b^{(l)}\) are the convolutional weights and bias, respectively, and \(p\) defines the radius of the convolution kernel, so that the full kernel size is \((2p+1) \times (2p+1)\). Here, \(\text{ReLU}(\cdot)\) denotes the Rectified Linear Unit activation function \citep{24}.   
    \item \textbf{Flattening:} The resulting spatial feature maps are flattened into a one-dimensional vector.
    \item \textbf{LSTM block:} Temporal dependencies are modelled using LSTM cells. At each time step $t$, the following equations are computed:
    \begin{equation}i_t = \sigma\left(W_i x_t + U_i h_{t-1} + b_i\right)\end{equation}
    \begin{equation}f_t = \sigma\left(W_f x_t + U_f h_{t-1} + b_f\right)\end{equation}
    \begin{equation}o_t = \sigma\left(W_o x_t + U_o h_{t-1} + b_o\right)\end{equation}
    \begin{equation}c_t = f_t \odot c_{t-1} + i_t \odot \tanh\left(W_c x_t + U_c h_{t-1} + b_c\right)\end{equation}
    \begin{equation}h_t = o_t \odot \tanh(c_t)\end{equation}
    where $W_{\{\cdot\}}$, $U_{\{\cdot\}}$, and $b_{\{\cdot\}}$ are learnable parameters and $\odot$ denotes element-wise multiplication.
    \item \textbf{Output layer:} A fully connected layer maps the LSTM output to produce the high-resolution projection.
\end{enumerate}

\item \textbf{ConvLSTM:} This model integrates two-dimensional convolution operations within the LSTM gates to preserve spatial structure while modelling temporal dynamics \citep{25}. at each time step \(t\), the ConvLSTM cell computes:
    \begin{equation}i_t = \sigma\Bigl(W_i * X_t + U_i * h_{t-1} + b_i\Bigr)\end{equation}
    \begin{equation}f_t = \sigma\Bigl(W_f * X_t + U_f * h_{t-1} + b_f\Bigr)\end{equation}
    \begin{equation}o_t = \sigma\Bigl(W_o * X_t + U_o * h_{t-1} + b_o\Bigr)\end{equation}
    \begin{equation}c_t = f_t \odot c_{t-1} + i_t \odot \tanh\Bigl(W_c * X_t + U_c * h_{t-1} + b_c\Bigr)\end{equation}
    \begin{equation}h_t = o_t \odot \tanh(c_t)\end{equation}
    where \( * \) denotes 2D convolution over the spatial dimensions. In this architecture, multiple stacked ConvLSTM layers are employed, batch normalization is applied after each convolution to stabilize training, and an up-sampling module is integrated to achieve the target resolution.
    
\item \textbf {ViT:} The model divides the input climate grid into patches and applies self-attention mechanisms to capture long-range spatial dependencies.
\begin{enumerate}
    \item \textbf{Patch embedding:} The input grid is partitioned into $N$ fixed-size patches. Each patch $X_i$ is flattened and projected linearly:
    \begin{equation}Z_0 = \begin{bmatrix} X_1E \\ X_2E \\ \vdots \\ X_NE \end{bmatrix} + E_{\text{pos}}\end{equation}
    where $E \in \mathbb{R}^{(p \times p \times c) \times d}$ is the learnable projection matrix and $E_{\text{pos}}$ provides positional encoding.
    \item \textbf{Transformer Encoder:} The sequence of patch embeddings is processed by a transformer encoder. The self-attention mechanism is given by:
    \begin{equation}\text{Attention}(Q,K,V) = \text{softmax}\!\left(\frac{QK^T}{\sqrt{d_k}}\right)V\end{equation}
    where $Q$, $K$, and $V$ denote the query, key, and value matrices, and $d_k$ is the key dimension.
    \item \textbf{Regression Head:} The encoder output is passed through a fully connected regression head to generate the high-resolution output.
\end{enumerate}

\item \textbf{GeoSTANet:} This model extends the ViT architecture by integrating geospatial and temporal encoding to capture the spatiotemporal variability in climate data. GeoSTANet specifically incorporates the geographic coordinates of each patch to increase spatial context, and it analyses the generated sequence along the temporal dimension. Each image patch is associated with geospatial coordinates - latitude and longitude. These coordinates are embedded into a higher-dimensional space using a learnable projection matrix \(W_{\text{geo}}\). Specifically, given a 2D coordinate vector \(X_{\text{latlon}} \in \mathbb{R}^2\), the geospatial encoding is computed as:
    \begin{equation}\text{GeoEnc} = W_{\text{geo}}\, X_{\text{latlon}}\end{equation}
    Where \(W_{\text{geo}} \in \mathbb{R}^{d \times 2}\) is a learnable matrix that projects the 2-dimensional coordinates into a \(d\)-dimensional embedding. This geospatial encoding is then combined (e.g., via concatenation or addition) with the corresponding patch embedding to provide explicit spatial context.
\begin{enumerate}     
    \item \textbf{Temporal transformer:} To model the temporal dynamics of the data, the sequence of enriched patch embeddings is processed by a dedicated transformer encoder that operates along the temporal dimension. Let \(H_t\) denote the hidden state at time \(t\). The temporal evolution is defined as:
    \begin{equation}H_t = \text{TransformerEncoder}(H_{t-1}), \quad \text{for } t \geq 1\end{equation}
    With the initial state \(H_0\) set as the geospatial enriched embedding from the first time step. This block enables the model to capture sequential dependencies over time, integrating both visual and geospatial information.   
    \item \textbf{Upsampling:} After the temporal processing, the final feature representation is passed through an upsampling module—such as a transposed convolution layer—to reconstruct the output on a high-resolution grid. This step is essential for applications like climate forecasting where detailed spatial predictions are required.
    \item \textbf{Input and output:} GeoSTANet accepts as input a time-ordered sequence of image patches, each accompanied by its geospatial coordinates. Initially, each patch is embedded using the standard ViT patch embedding method. The geospatial coordinates are then projected via \(W_{\text{geo}}\) into a \(d\)-dimensional space and combined with the patch embeddings. The resulting sequence is processed by the temporal transformer block to model the temporal dependencies, and finally, the features are up-sampled to produce a high-resolution output. This pipeline distinguishes GeoSTANet from the previously described CNN and ConvLSTM models by explicitly incorporating geographic context and dedicated temporal processing.
\end{enumerate}

\end{enumerate}

This complete approach combines advanced deep-learning-based downscaling systems with a data-driven DL-TOPSIS ranking algorithm. By combining multi-criteria model evaluation with robust spatial and temporal feature extraction, the framework produces high-resolution climate projections that accurately capture both mean behaviour and extremes, thereby supporting informed regional climate impact assessments.

\section{Results and discussions}

The reported results cover the performance of model ranking via the proposed DL-TOPSIS framework and the statistical downscaling via the four deep-learning models discussed above. Starting with coarse-resolution CMIP6 model evaluation, these results offer a whole picture of model fidelity at several levels, and then deep-learning transforms top-ranked GCM outputs into high-resolution fields. This approach yields model strengths and limits in various seasons and climate zones as well as measuring gains made by the downscaling process. 

\subsection{Model ranking through DL-TOPSIS}

\begin{landscape}
\begin{figure}[h]
    \centering
    \includegraphics[width=0.95\paperwidth, height=0.95\paperheight, keepaspectratio]{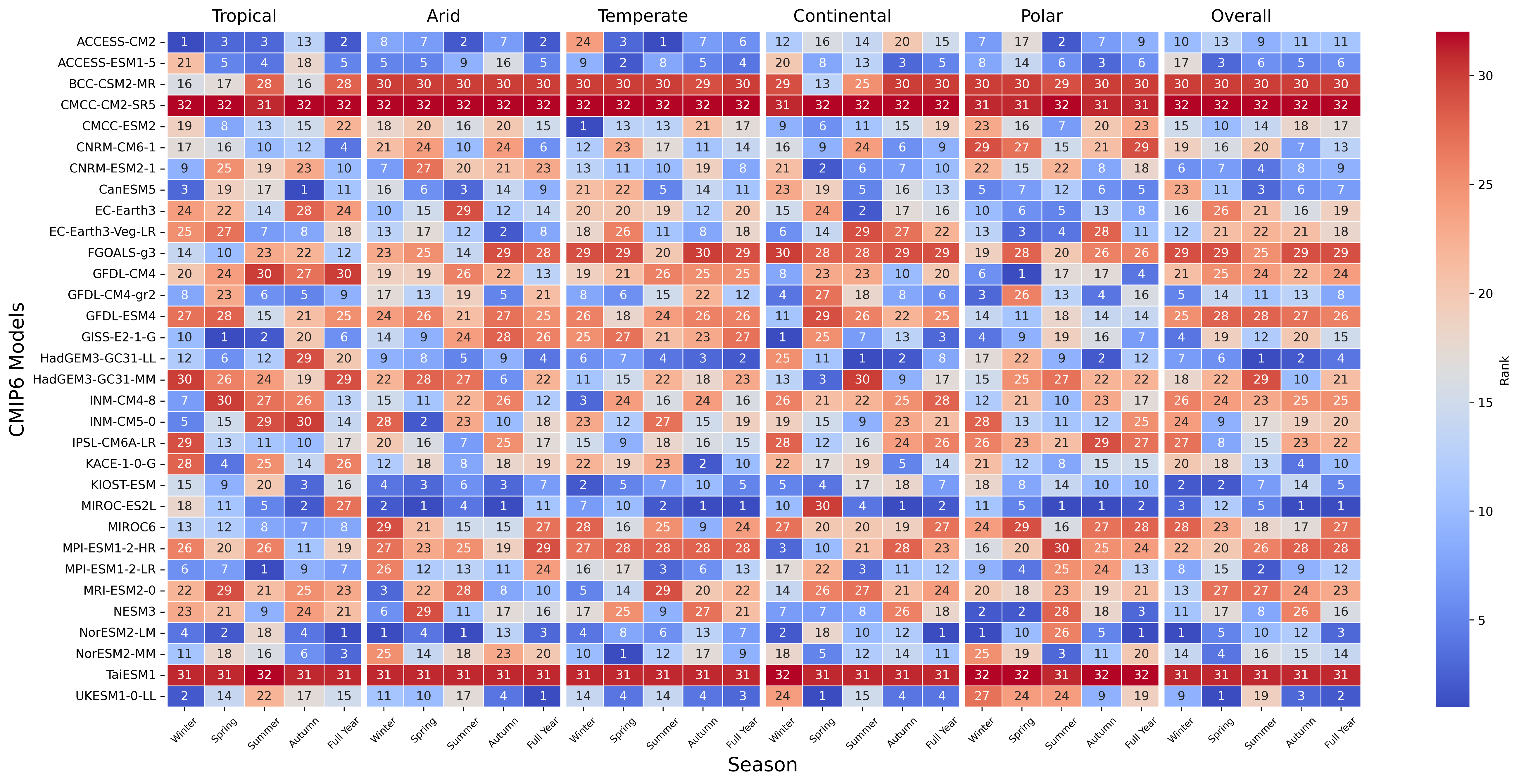}
    \caption{CMIP6 model ranking heatmap across European climate zones and seasons.}
    \label{fig:3}
\end{figure}
\end{landscape}

The DL-TOPSIS approach was used to do a complete ranking of the thirty-two CMIP6 models. These ranking results for every climate zone (Tropical, Arid, Temperate, Continental, and Polar) spanning winter, spring, summer, autumn, and the whole year are shown on the heat map in Figure \ref{fig:3}. Better ratings are indicated by cooler (blue) tones; poorer ratings by warmer (red) tones. Particularly in temperate and continental areas, which are difficult due to great seasonal variability and complex precipitation regimes, NorESM2-LM, HadGEM3-GC31-LL, and GISS-E2-1-G routinely ranked among the best models across many zones. Among the lowest-ranked models were TaiESM1, CMCC-CM2-SR5, BCC-CSM2-MR, FGOALS-g3, and showing notable temperature mean representation biases and errors. While many models battled with cold extremes, NorESM2-LM and MPI-ESM1-2-LR excelled others in showing subzero temperature ranges (TNn) in Polar zones.

Table \ref{tab:3}, represents the top 5 CMIP6 Models for each season and over Europe, Using performance evaluation criteria such as Score, Bias, RMSE, KGE, NSE, and PDF Overlap the table provides a seasonal evaluation of CMIP6 models throughout six climate zones (Tropical, Arid, Temperate, Continental, Polar, and Overall). Especially in the Polar region, Winter suggests NorESM2-LM performs better in many zones with the best RMSE. GISS-E2-1-G leads in the Tropical zone, Spring denotes UKESM1-0-LL as the top performer. Summer shows HadGEM3-GC31-LL, and MPI-ESM1-2-LR perform well in the Tropical zone. In all zones, MIROC-ES2L does well in Autumn showing great PDF Overlap and NSE. Over the Full Year, MIROC-ES2L turns out to be the best overall model, and NorESM2-LM performs well in the Polar and Continental zones. Although higher KGE and NSE show a stronger correlation with observations, bias and RMSE trends indicate better accuracy from lower levels. The results highlight seasonal changes in model performance. These multi-model assessments are important since they expose no one universal model that performs under all circumstances. Rather, the top five can be seen as a group of good candidates, each with particular characteristics (e.g., strong skill in heat extremes or cold extremes) that can be used for different climate-sensitive applications.

\begin{landscape}
\renewcommand{\arraystretch}{0.95}
\setlength{\tabcolsep}{8pt}
\begin{longtable}{l l l c c c c c c}
    \caption{Top 5 CMIP6 models by climate zone and season.} \label{tab:3} \\
    \toprule
    \textbf{Season} & \textbf{Zone} & \textbf{Model} & \textbf{Score} & \textbf{Bias} & \textbf{RMSE} & \textbf{KGE} & \textbf{NSE} & \textbf{PDF Overlap} \\
    \midrule
    \endfirsthead
    \toprule
    \textbf{Season} & \textbf{Zone} & \textbf{Model} & \textbf{Score} & \textbf{Bias} & \textbf{RMSE} & \textbf{KGE} & \textbf{NSE} & \textbf{PDF Overlap} \\
    \midrule
    \endhead
    \midrule
    \multicolumn{9}{r}{\textit{Continued on next page}} \\
    \midrule
    \endfoot
    \bottomrule
    \endlastfoot
    \multirow{6}{*}{Winter (DJF)} & Tropical & ACCESS-CM2 & 0.9501 & 0.3596 & 4.6423 & 0.0585 & \textbf{-0.8327} & 0.6908 \\
    & Arid & NorESM2-LM & 0.9333 & -0.4068 & 5.0143 & 0.1041 & -0.8527 & \textbf{0.9626} \\
    & Temperate & CMCC-ESM2 & 0.9277 & -0.0098 & 5.1524 & 0.0817 & -1.5013 & 0.8205 \\
    & Continental & GISS-E2-1-G & 0.9103 & 0.3599 & 6.6133 & 0.0913 & -2.0429 & 0.8450 \\
    & Polar & NorESM2-LM & \textbf{0.9542} & -1.8734 & \textbf{4.3212} & 0.0307 & -1.5815 & 0.4112 \\
    & Overall & NorESM2-LM & 0.9234 & \textbf{0.0008} & 5.7904 & \textbf{0.1376} & -1.3970 & 0.8968 \\    
    \midrule
    \multirow{6}{*}{Spring (MAM)} & Tropical & GISS-E2-1-G & \textbf{0.9335} & 0.8253 & 5.1894 & 0.1516 & \textbf{-0.5558} & \textbf{0.8763} \\
    & Arid & MIROC-ES2L & 0.9142 & -0.6355 & 5.1720 & 0.1324 & -0.5849 & 0.8500 \\
    & Temperate & NorESM2-MM & 0.9305 & -0.2988 & 5.4334 & 0.1444 & -0.6687 & 0.7891 \\
    & Continental & UKESM1-0-LL & 0.9253 & \textbf{-0.0932} & 6.0145 & 0.1288 & -0.9561 & 0.8234 \\
    & Polar & GFDL-CM4 & 0.9042 & -1.1415 & \textbf{3.9518} & 0.0189 & -1.3881 & 0.4376 \\
    & Overall & UKESM1-0-LL & 0.9286 & -0.2803 & 5.6790 & \textbf{0.1821} & -0.7016 & 0.8384 \\    
    \midrule
    \multirow{6}{*}{Summer (JJA)} & Tropical & MPI-ESM1-2-LR & \textbf{0.9429} & 1.1819 & 4.8028 & 0.2363 & \textbf{-0.2472} & 0.7864 \\
    & Arid & NorESM2-LM & 0.8794 & -0.1021 & 4.4709 & 0.2579 & -0.5831 & 0.7724 \\
    & Temperate & ACCESS-CM2 & 0.9189 & -0.3081 & 5.2273 & 0.0928 & -0.9402 & 0.8255 \\
    & Continental & HadGEM3-GC31-LL & 0.9131 & 0.2221 & 4.8654 & 0.1478 & -0.7857 & 0.8489 \\
    & Polar & MIROC-ES2L & 0.9235 & -1.0053 & \textbf{2.7864} & 0.0616 & -0.6326 & 0.4830 \\
    & Overall & HadGEM3-GC31-LL & 0.9322 & \textbf{0.0261} & 4.9257 & \textbf{0.2812} & -0.5076 & \textbf{0.8623} \\
    \midrule
    \multirow{6}{*}{Autumn (SON)} & Tropical & CanESM5 & 0.9324 & 1.1847 & 4.7562 & 0.3193 & -0.2525 & 0.8518 \\
    & Arid & MIROC-ES2L & 0.9096 & -0.2656 & 5.0993 & 0.3271 & -0.2698 & 0.8453 \\
    & Temperate & MIROC-ES2L & 0.9097 & -0.2064 & 4.7253 & 0.2887 & -0.3334 & 0.8233 \\
    & Continental & MIROC-ES2L & 0.9257 & \textbf{-0.0333} & 4.3083 & 0.2709 & -0.4697 & \textbf{0.9152} \\
    & Polar & MIROC-ES2L & 0.9412 & -1.7524 & \textbf{3.4698} & 0.0641 & -1.2028 & 0.3976 \\
    & Overall & MIROC-ES2L & \textbf{0.9502} & -0.1184 & 4.5932 & \textbf{0.4399} & \textbf{-0.0892} & 0.8851 \\
    \midrule
    \multirow{6}{*}{Full Year} & Tropical & NorESM2-LM & 0.9192 & 0.8440 & 4.8845 & 0.3179 & \textbf{-0.2764} & 0.8386 \\
    & Arid & UKESM1-0-LL & 0.9201 & -0.2241 & 5.1109 & 0.3529 & -0.3425 & \textbf{0.8980} \\
    & Temperate & MIROC-ES2L & 0.9105 & -0.2939 & 4.9621 & 0.3198 & -0.3238 & 0.8455 \\
    & Continental & NorESM2-LM & 0.9225 & 0.1673 & 5.5920 & 0.2487 & -0.8120 & 0.8843 \\
    & Polar & NorESM2-LM & \textbf{0.9408} & -1.5537 & \textbf{3.7186} & 0.0498 & -1.2792 & 0.4842 \\
    & Overall & MIROC-ES2L & 0.9312 & \textbf{-0.1362} & 5.0937 & \textbf{0.3607} & -0.3015 & 0.8934 \\
\end{longtable}
\end{landscape}

\begin{figure}[h]
    \centering
    \includegraphics[width=0.8\textwidth]{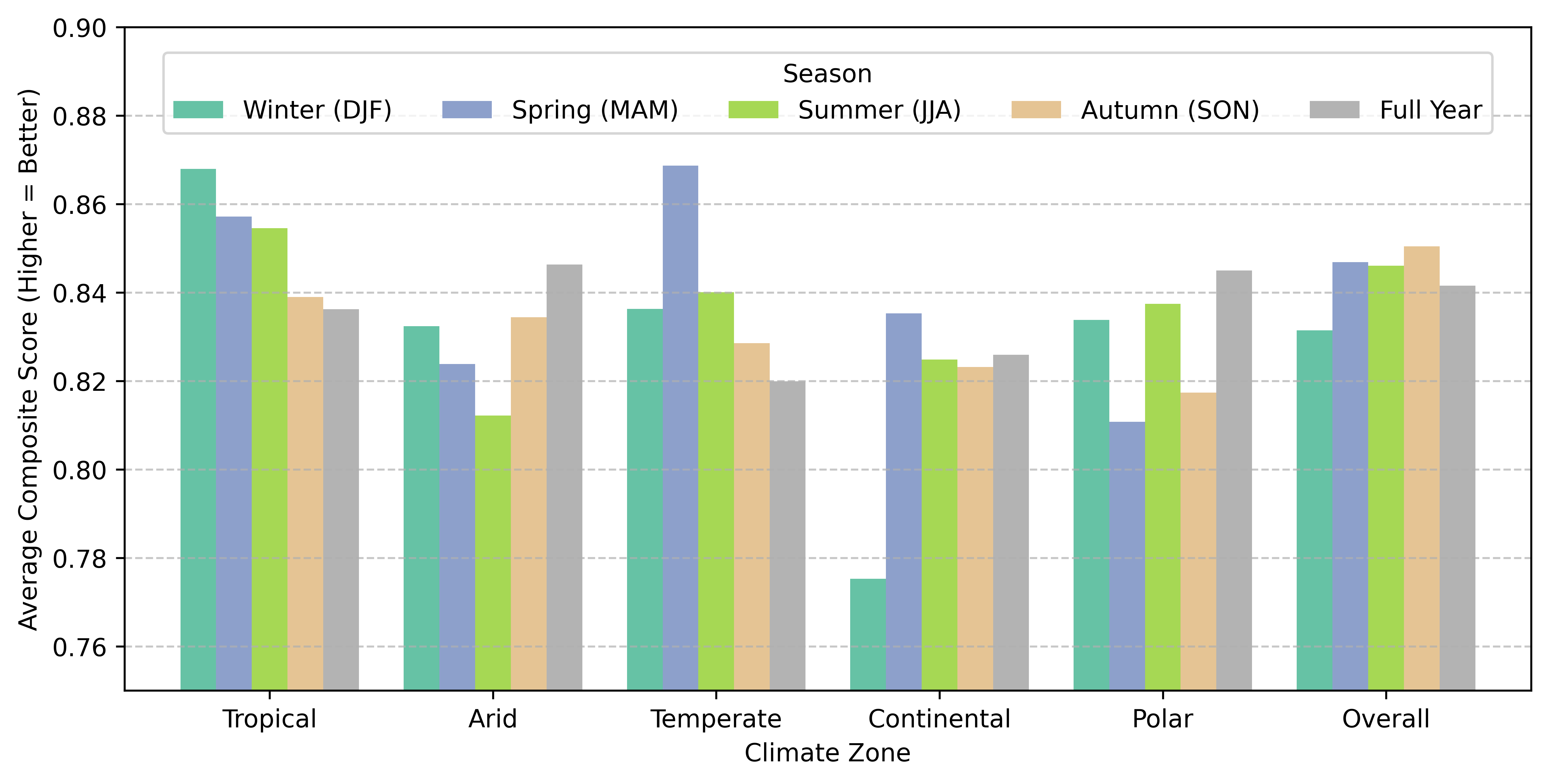}
    \caption{Overall Performance of CMIP6 Models in Europe.}
    \label{fig:4}
\end{figure}

\begin{figure}[h]
    \centering
    \includegraphics[width=0.7\paperwidth, height=0.7\paperheight, keepaspectratio]{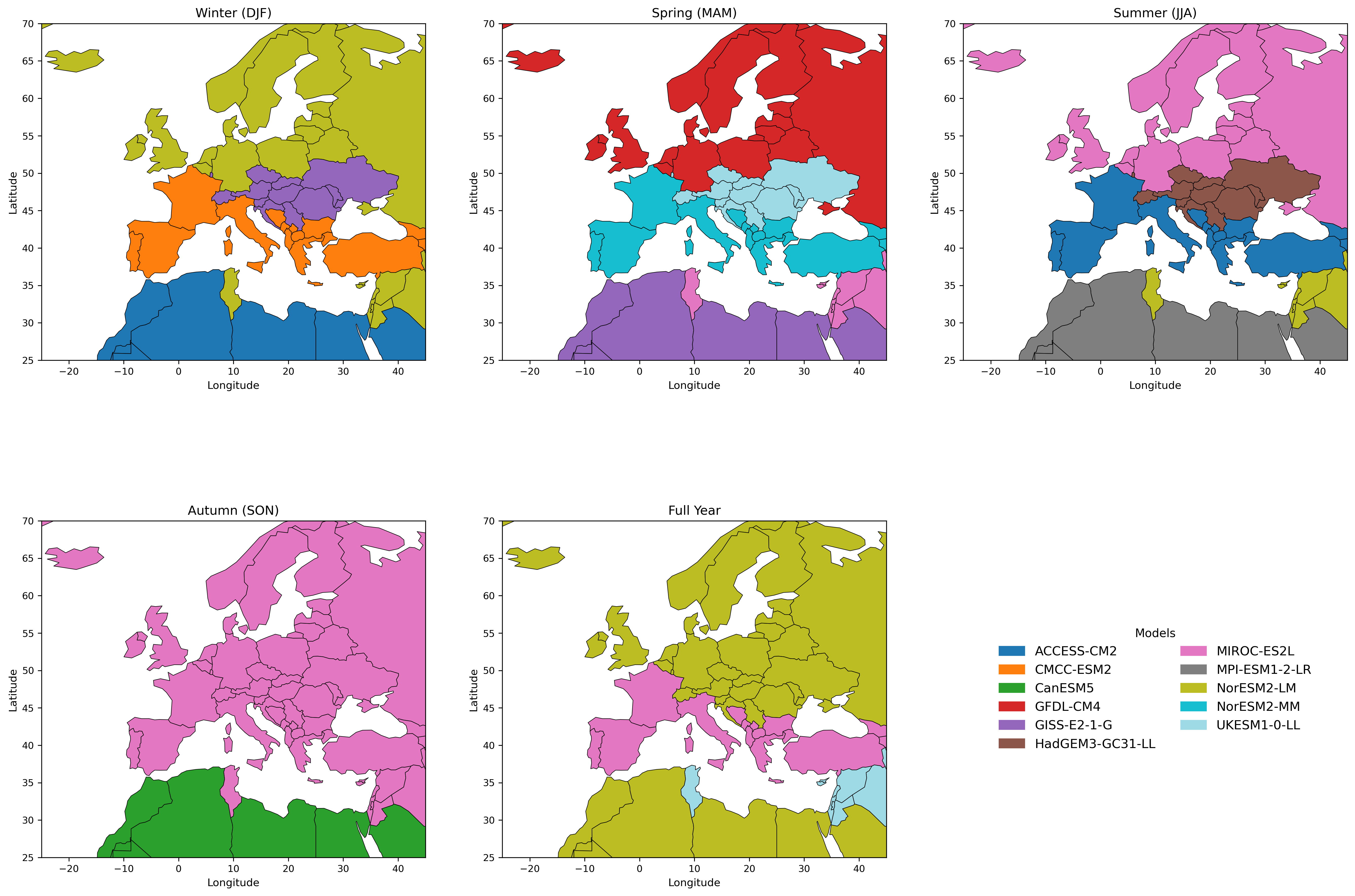}
    \caption{Spatial distribution of top-ranked CMIP6 models in Europe.}
    \label{fig:5}
\end{figure}

Figure \ref{fig:4} illustrates the overall CMIP6 GCMs Model performance by climate zones and seasons. The plot shows the CMIP6 GCMs' average performance ratings across several climate zones and seasons, ranging from 0.775 to 0.868. A higher score indicates a better match with the actual observation. Temperate and tropical zones exhibit optimal conditions in the spring and winter, indicating that models adequately represent seasonal fluctuations in these regions (scores >  0.86). In contrast, the Continental zone has the poorest Winter performance (0.775), indicating difficulties in modelling cold-season climate processes. The Overall category shows model endurance by maintaining consistent performance (0.84-0.85) across seasons. The Polar zone fluctuates; it peaks in the Full Year (0.844) but drops in the Spring (0.810), indicating difficulties in maintaining high-latitude activity. The arid zone is stable (~0.81-0.83) with minimal seasonal impact. These findings highlight the importance of model evaluations based on regional and seasonal contexts, as they show that climate models are generally more reliable in certain seasons and locations. Figure \ref{fig:5} displays the best-ranked CMIP6 GCM spatially across Europe over various Köppen-Geiger Climate Zones. 

\subsection{Downscaling performance results}

\begin{landscape}
\begin{figure}[h]
    \centering
    \includegraphics[width=0.99\paperwidth, height=0.99\paperheight, keepaspectratio]{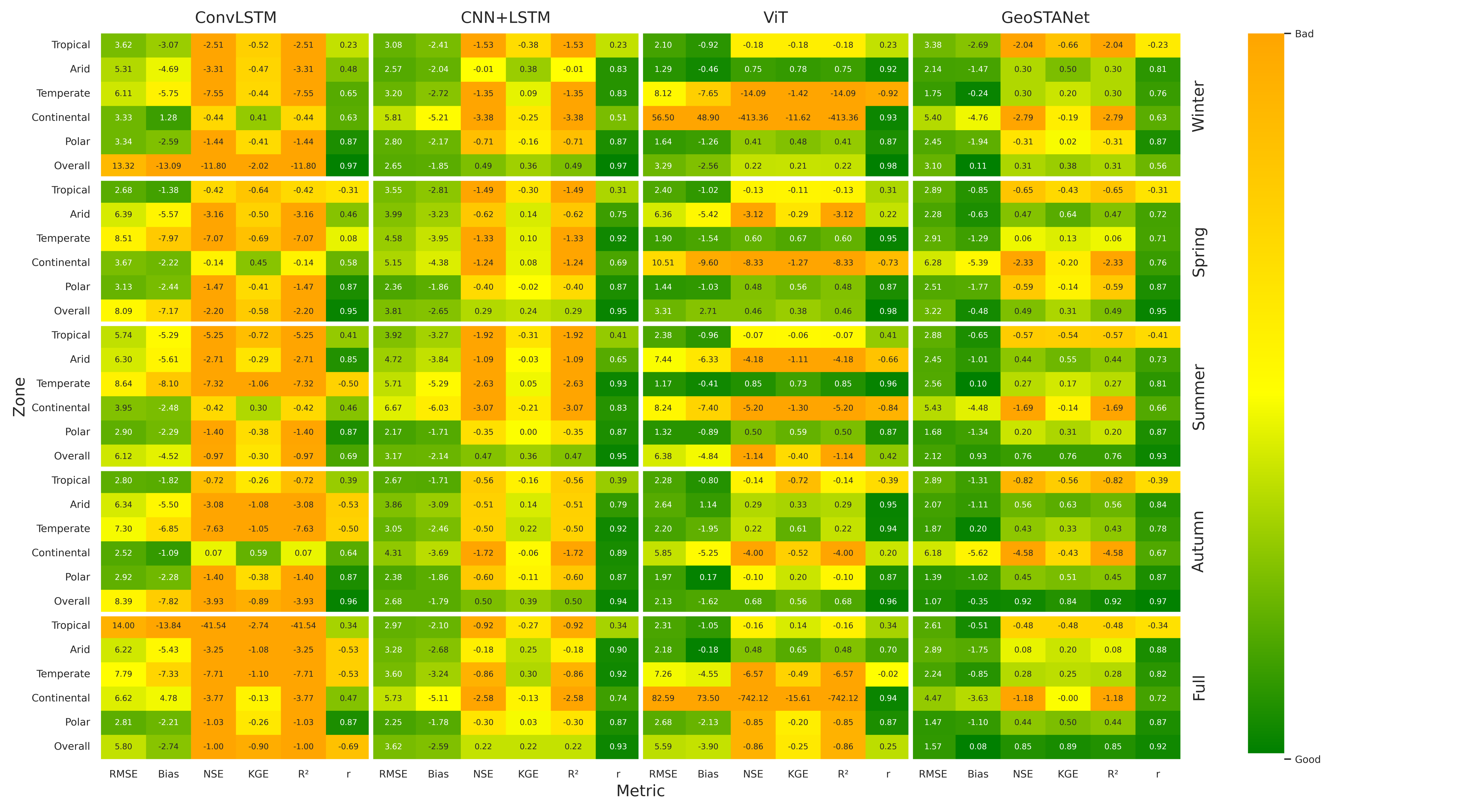}
    \caption{Performance comparison of downscaling models across zones and seasons.}
    \label{fig:6}
\end{figure}
\end{landscape}

After determining the best-performing models for every season and zone, four advanced deep-learning architectures CNN-LSTM, ConvLSTM, ViT, and GeoSTANet downscaled the selected GCM outputs to a fine-resolution grid (0.1°×0.1°). Across the same temperature zones and seasons, Figure \ref{fig:6} offers a relative visualization of different structures. Deeper greens indicate better competence in performance measures like Bias, RMSE, NSE, KGE, correlation (r), and PDF overlap—which are color-coded. Especially in capturing DTR and extremes TXx and TNn, this picture emphasizes how each design manages the spatiotemporal complexity of daily temperature fields. Figure \ref{fig:6} represents a heat map of performance evaluation metrics of downscaling for various zones and seasons over Europe using various deep-learning models.

In areas with moderate geographical variability—that is, temperate and continental zones, CNN-LSTM showed particularly high accuracy. While the LSTM units modelled daily-to-season temperature variations, the convolutional layers efficiently identified spatially localized characteristics. CNN-LSTM did, however, occasionally show over-smoothing in locations with steep gradients, such as coastal zones or mountainous areas, clearly shown in somewhat higher RMSE values. In areas with fast temperature transitions e.g., mountainous borders between continental and polar zones - ConvLSTM sustained spatial coherence better than CNN-LSTM, by incorporating convolution operations directly into the LSTM gating mechanism. This benefit was particularly evident in winter when daily maximum and minimum temperatures can be quite influenced by convective processes. ConvLSTM did, however, occasionally show training stability problems that needed careful learning rate and batch size tweaking to prevent over-fitting.

The ViT model downscaled the results by treating each climate map as a set of patches and using multi-head self-attention. Particularly in tropical and subtropical zones, where broad-scale circulation can control temperature distributions, this method was quite good in catching large-scale atmospheric patterns and interconnections. ViT did, however, sometimes suffer from local topographic effects since self-attention may not always prioritize fine-scale terrain characteristics unless the patch size and positional embeddings are precisely optimized. ViT yielded competitive KGE and correlation values in arid zones; small-scale extremes occasionally seemed unnaturally smoothed, suggesting that patch-based embeddings may need more fine-grasping to manage localized events.

Particularly in arid zones and TNn in polar zones, GeoSTANet, the geospatial-spatiotemporal transformer, routinely outperformed the other architectures in capturing temperature extremes. GeoSTANet dynamically learned which areas and time steps were most important for forecasting daily maxima and minima by including explicit geographical encoding (latitude and longitude embeddings) and temporal attention blocks. In demanding environments—including the transitional zones between temperate and arctic climates—this capacity produced reduced bias and RMSE values. Higher PDF overlap scores in some zones suggest that GeoSTANet was able to faithfully replicate the distribution tails for daily temperature using synergy between attention-based mechanisms and an imbalance-aware training method (focusing on rare extremes). For climate impact studies, which usually rely on accurate forecasts of unusual events like heat waves or severe cold spells, such an advantage is highly desired.

Downscaling accuracy is significantly impacted by the interaction of the GCM choice with downscaling architecture. Using highly ranked GCMs with low bias and variability minimizes the correctional load on statistical downscalers, thus lowering residual errors. On the other hand, even the most sophisticated downscaling model inherits large-scale biases when associated with poorly ranked GCMs including CMCC-CM2-SR5, TaiESM1, and BCC-CSM2-MR, which displayed systematic biases and high RMSE across many climate zones. Particularly in the Continental and Temperate zones, where temperature variability is high, our analysis shows that pairing top-performing GCMs—NorESM2-LM, GISS-E2-1-G, and HadGEM3-GC31-LL—with GeoSTANet routinely performs better than other combinations across seasonal and annual scales. Maintaining strong KGE and NSE values, this combination achieves RMSE cuts of up to 20\% over the next-best alternative. These data highlight the need for a two-stage strategy: a) Strong multi-metric evaluation of GCMs to choose the most dependable climate forecasts. b) Using GeoSTANet, advanced deep-learning-based downscaling captures spatiotemporal dependencies and sub-grid processes, guaranteeing high-resolution climate forecasts.

\subsection{Seasonal performance results}

Seasonal assessments of the downscaled outputs show important variations in the capacity of every architecture to replicate temperature extremes. Especially in continental and dry zones, convective and radiative processes produce significant diurnal variations throughout summer. While ConvLSTM performed better, CNN-LSTM periodically under-predicted TXx, presumably because the convolutional gating preserved local convective fingerprints. Although ViT's patch-based method usually performed well in capturing more general patterns, it may have missed small-scale heat islands, which would have somewhat understated the peak daily maximum. Higher correlation values and better PDF overlap for TXx distributions show GeoSTANet's most skilful resolution of these localized hot spots. Errors in winter tended to gather around significant temperature inversions or cold extremes. While GeoSTANet once more excelled by using temporal attention to track the development of cold air masses, ConvLSTM controlled spatiotemporal transitions well. In particular, biases were usually smaller in winter than in summer, implying that downscaling designs may find simpler large-scale synoptic circumstances to record.

\subsubsection{Regional performance results}

From a regional standpoint, the desert zone (B) presented special difficulties because of sharp daily fluctuations; the polar zone (E) necessitated strong handling of negative temperature extremes. CNN-LSTM and ConvLSTM both demonstrated a rather good ability in arid zones to capture daily temperature fluctuations; nevertheless, if the GCM inputs included systematic warm or cold biases, they could overstate the magnitude of extremes. Advanced attention levels of GeoSTANet regularly reduced these biases, suggesting that attention-based designs are appropriate for arid areas with high radiative forcing. In polar areas, model evaluation proved much dependent on the ability to depict negative temperature extremes (TNn). While ConvLSTM performed better because of its inherent spatiotemporal gating, CNN-LSTM occasionally suffered with capturing extended cold spells if they were not prominent in the training process. In these high-latitude areas, GeoSTANet's geographic encoding was particularly helpful since it more closely matched observed data to temperature projections.
PDF overlap was utilized to assess not only mean values but also the distribution of temperature, therefore evaluating each architecture. GeoSTANet's better depiction of the whole temperature distribution, including both central trends and tails, clearly showed consistently higher overlap scores than the other models. For research on climate change, where precise tail behaviour can imply the difference between an underestimated or realistically expressed risk of extreme events, this advantage is essential. The performance of ViT in PDF overlap was partially reliant on patch size and training procedures; hence, it suggests possible improvements if patch embedding or positional encodings were optimized for tasks related to the climate. Although CNN-LSTM and ConvLSTM usually showed modest overlap scores, they occasionally revealed small changes in the distribution tails depending on the training data size or if the GCM inputs included persistent biases.

\subsection{Discussion and outlook}

The results naturally validate that advanced deep-learning architectures could significantly improve daily temperature data's spatial and temporal accuracy. The success of GeoSTANet highlights how important specialized design choices, geospatial encoding and attention-based mechanisms are to capture climate-related variability. Regardless, CNN-LSTM and ConvLSTM are competitive options, particularly in computationally limited environments, and can create downscaled fields that exceed conventional statistical approaches. ViT distinguishes itself for recording notable connections and patterns, even though it may need more fine-tuning for localized aspects. It is crucial to underline the wider implications for climate adaptation and decision-making even as one discusses these outcomes. Reliable high-quality temperature fields assist in enhancing risk assessments in public health, infrastructure design, and agriculture. For instance, whereas accurate modelling of TXx in desert and temperate zones can guide early warning systems for heat waves, a strong representation of TNn helps winter hazard planning and ecosystem preservation activities. Moreover, the suggested DL-TOPSIS structure takes advantage of the synergy between the downscaling technique and improved model choice to give decision-makers more consistent data. Instead of depending just on single GCM outputs or simpler downscaling methods, this two-tiered approach targets the most reliable global models and improves them with state-of-the-art neural architectures for finer detail.

Future studies should consider extending this approach to other variables such as precipitation, wind speed, or soil moisture, where the interaction of local effects and large-scale circulation may differ greatly from temperature fields. Further increasing confidence in the downscaled products could be ensemble-based methods including Bayesian uncertainty quantification or combining several deep-learning architectures. Employing several emission scenarios e.g., SSP1-2.6, SSP5-8.5 these architectures would also guide changes in model biases under future warming and whether sophisticated downscalers are still robust. Combining several observational or reanalysis products such as ground-based station networks or MERRA-2 may provide a more complete training and validation dataset, possibly improving model performance in data-sparse areas such as mountainous or high-latitude regions. Considering these, modern deep-learning models seem to be quite able to refine the coarse outputs of top-ranked CMIP6 GCMs. GeoSTANet shows to be the most consistent design across several climate zones and seasons especially in capturing distribution extremes TXx and TNn and obtaining high PDF overlap. Still, excellent choices are CNN-LSTM and ConvLSTM; ConvLSTM is especially good at preserving spatial coherence in regions with strong temperature gradients. ViT shows promise in gathering general climate characteristics but might need local-scale event-specific corrections. These findings confirm the need of exactly match strong GCMs with advanced downscaling models to produce high-resolution climate forecasts that regularly direct scientific research, policy, and adaptation strategies.

\section{Conclusions}

Combining a data-driven GCM ranking system (DL-TOPSIS) with deep-learning-based downscaling, this work presents a strong two-stage framework for high-resolution climate projections over Europe. Objectively evaluating 32 CMIP6 models across five Köppen-Geiger climate zones (Tropical, Arid, Temperate, Continental, and Polar) and several seasons (Winter, Spring, Summer, Autumn, and Full Year) the ranking system dynamically assigns weights to performance metrics to lower bias. While CMCC-CM2-SR5, TaiESM1, and BCC-CSM2-MR show systematic biases and weak correlation with observations, so less suitable for high-resolution downscaling, the results confirm that NorESm2-LM, GISS-E2-1-G, HadGEM3-GC31-LL, MPI-ESm1-2-LR, and ACCESS-CM2-SR5 consistently outperform other models in different climate conditions.

Four advanced deep-learning architectures—CNN-LSTM, CNN-LSTM, ViT, GeoSTANet—were used in the second stage to downscale top-ranked GCM outputs to a fine-scale resolution (0.1° × 0.1°). With a 20\% RMSE decrease, GeoSTANet stands out as the most successful downscaling model since it achieves statistically significant improvements over other methods. Extreme temperature fluctuations are faithfully captured by its geospatial and temporal attention mechanisms, preserving high KGE (0.89), NSE (0.85), and PDF overlap scores (0.91). Whereas CNN-LSTM improves temporal coherence, ConvLSTM also performs well in areas with fast spatial transitions. ViT needs more tuning to improve fine-scale resolution even if it shines in catching broad climatological trends.

These findings underline the need to combine advanced downscaling methods with ideal model selection to improve regional climate projections. Through better daily temperature forecasts, this framework offers insightful analysis for infrastructure design, climate risk assessments, and adaptation strategies. Future research will concentrate on extending this framework using multi-variable downscaling architectures to other important climate variables including precipitation extremes, wind fields, and soil moisture. We also wish to evaluate model confidence levels by including Bayesian uncertainty quantification. We will also discuss the generalization of this method to other geographical areas, including East Asia and North America. At last, ensemble-based approaches will be examined to increase resilience under several emission scenarios (SSP1-2.6, SSP5-8.5).

\section*{Declarations}

\textbf{Acknowledgements} The authors acknowledge the SESAR 3 Joint Undertaking and its members for their support in funding this research under grant agreement No. 101114795 as part of the E-CONTRAIL project. We also appreciate ECMWF for providing the ERA5 reanalysis dataset and WCRP for facilitating CMIP6 data access. Special thanks to our colleagues and collaborators for their valuable insights.

\textbf{Funding} This research was funded by the SESAR 3 Joint Undertaking under the E-CONTRAIL project (Grant Agreement No. 101114795).

\textbf{Authors’ Contributions} Parthiban Loganathan: Conceptualization, data collection, investigation, writing, and visualization. Elias Zea, Ricardo Vinuesa, and Evelyn Otero: Project administration, review, and editing.

\textbf{Ethical Approval} Not applicable.

\textbf{Consent to Participate} Not applicable.

\textbf{Consent to Publish} All authors consent to the publication of this manuscript.

\textbf{Competing Interests} The authors declare no competing interests.

\textbf{Data Availability Statement} The data used in this study is included in the manuscript. Additional data or supplementary materials can be provided upon reasonable request.

\bibliography{ref}
\end{document}